\documentclass[11pt]{article}

\usepackage[final]{acl}

\usepackage{times}
\usepackage{latexsym}

\usepackage[T1]{fontenc}

\usepackage[utf8]{inputenc}

\usepackage{hyperref}
\usepackage{footmisc}
\usepackage{setspace}
\usepackage[nopatch=footnote]{microtype}

\usepackage{inconsolata}

\usepackage{url}
\usepackage{enumitem}
\usepackage{graphicx}
\usepackage[table]{xcolor}
\usepackage{xcolor}
\usepackage{threeparttable}
\usepackage{framed}
\usepackage{algorithm}
\usepackage{algpseudocode}
\usepackage{multirow}
\usepackage{pifont}
\usepackage{amsmath}
\usepackage{amssymb} 
\usepackage{fvextra}
\usepackage{listings}
\usepackage{booktabs}
\usepackage[most]{tcolorbox}
\usepackage{lipsum} 

\newcommand{\simop}{\text{sim}}
\newcommand{\rank}{\text{rank}}
\newcommand{\Beta}{\text{Beta}}
\newcommand{\LI}{\text{LI}}
\title{MAB-DQA: Addressing Query Aspect Importance in Document Question Answering with Multi-Armed Bandits}

\author{
 \textbf{Yixin Xiang\textsuperscript{1}},
 \textbf{Yunshan Ma\textsuperscript{2}},
 \textbf{Xiaoyu Du\textsuperscript{1}\thanks{\noindent Corresponding author.}},
 \textbf{Yibing Chen\textsuperscript{3}},
 \textbf{Yanxin Zhang\textsuperscript{4}},
 \textbf{Jinhui Tang\textsuperscript{5}},
\\
 \textsuperscript{1}Nanjing University of Science and Technology,\\
 \textsuperscript{2}Singapore Management University,
 \textsuperscript{3}Nanjing Pami Intelligent Technology Co., Ltd.,\\
 \textsuperscript{4}University of Wisconsin - Madison,
 \textsuperscript{5}Nanjing Forestry University
\\
\texttt{elephantoh@qq.com, ysma@smu.edu.sg, duxy@njust.edu.cn,}
\\
\texttt{chenyibing@pami-ai.com, yzhang2879@wisc.edu, tangjh@njfu.edu.cn}
}

\begin{document}
\maketitle

\begin{abstract}
Document Question Answering (DQA) involves generating answers from a document based on a user's query, representing a key task in document understanding. This task requires interpreting visual layouts, which has prompted recent studies to adopt multimodal Retrieval-Augmented Generation (RAG) that processes page images for answer generation. However, in multimodal RAG, visual DQA struggles to utilize a large number of images effectively, as the retrieval stage often retains only a few candidate pages (e.g., Top-4), causing informative but less visually salient content to be overlooked in favor of common yet low-information pages. To address this issue, we propose a Multi-Armed Bandit-based DQA framework (MAB-DQA) to explicitly model the varying importance of multiple implicit aspects in a query. Specifically, MAB-DQA decomposes a query into aspect-aware subqueries and retrieves an aspect-specific candidate set for each. It treats each subquery as an arm and uses preliminary reasoning results from a small number of representative pages as reward signals to estimate aspect utility. Guided by an exploration-exploitation policy, MAB-DQA dynamically reallocates retrieval budgets toward high-value aspects. With the most informative pages and their correlations, MAB-DQA generates the expected results. On four benchmarks, MAB-DQA shows an average improvement of 5\%-18\% over the state-of-the-art method, consistently enhancing document understanding.
Codes are available at \url{https://github.com/ElephantOH/MAB-DQA}.
\end{abstract}
\section{Introduction}
\label{introduction}

\begin{figure}[ht]
\centering
\includegraphics[width=1\linewidth]{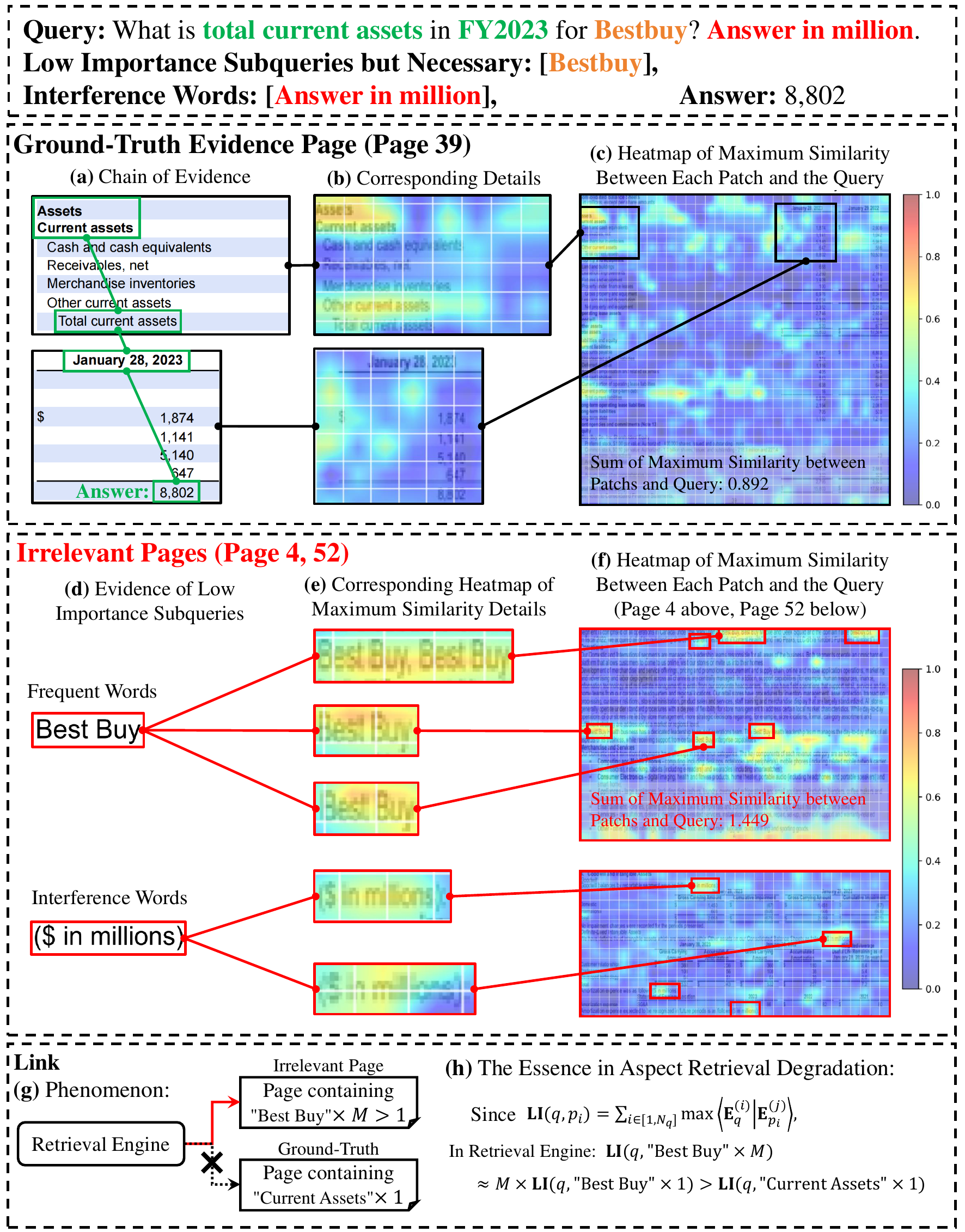}
\caption{
Aspect Retrieval Degradation in DQA. 
(a) Ground-truth evidence and answer.
(b-c) Similarity heatmap between query and patches.
(d) Aspects of low importance in the question.
(e-f) High spurious similarity on irrelevant pages due to frequent terms.
(g-h) Aspects of low importance' aggregated score can exceed crucial evidence.}
\label{fig:photo}
\end{figure}

Document Question Answering requires AI to answer user questions about given documents~\cite{slidevqa, acl:qa:1}.
DQA proves valuable in real‑world applications such as financial forms, medical report interpretation, and academic literature assistance~\cite{ye_boosting_2024, acl:qa:4}.
Accomplishing this task relies on visual‑language models and retrieval‑augmented generation (RAG) to understand long documents with complex layouts~\cite{zhou_vista_2024}.
Existing advanced approaches, such as Colpali~\citep{colpali} and MoloRAG~\citep{molo_rag}, adopt a vision‑query late interaction (LI) paradigm for retrieval.
They compute the dot product between each query token embedding and all document image patch embeddings, retaining the maximum similarity score between each query token and the most relevant image patch~\cite{colbert}.
This operator preserves fine‑grained token‑patch interactions but merely performs a mechanical "max‑pooling + summation" operation.
As a result, it cannot mimic the human ability to weigh the importance of multiple aspects in a query and selectively focus on them, which limits performance on complex multi‑aspect questions.

Fig.~\ref{fig:photo} illustrates an example from MMLongBench using a financial report.
Fig.~\ref{fig:photo}(a)‑(c) show the maximum similarity scores obtained by Colpali on the ground‑truth evidence page (Page 39).
The green box in Fig.~\ref{fig:photo}(a) highlights the correct retrieval chain, while Fig.~\ref{fig:photo}(b)‑(c) display interpretability heatmaps.
The color in the heatmaps indicates the maximum similarity (dot product) between each image patch embedding and the query, with red representing high similarity and blue indicating irrelevance.
Based on the LI paradigm, Colpali assigns Page 39 a score of $\textbf{LI}(\text{Query}, \text{Page 39})=0.892$, as shown in Fig.~\ref{fig:photo}(c).
However, as seen in Fig.~\ref{fig:photo}(d), pages unrelated to the answer, such as those containing high‑frequency but low‑importance words like "Best Buy" or irrelevant terms like "million", can also receive high scores (Fig.~\ref{fig:photo}(e)‑(f)), such as $\textbf{LI}(\text{Query}, \text{Page 4})=1.449$. 

To address this limitation, we propose the MAB‑DQA, a multi‑armed bandit-guided DQA framework that explicitly models and exploits the varying importance of multiple implicit aspects in a query. Our core idea is to dynamically decompose a query into aspect‑aware subqueries, treat each subquery as an arm in a multi‑armed bandit, and use preliminary reasoning feedback as a reward signal to estimate the utility of each aspect. Guided by an exploration-exploitation policy, MAB‑DQA reallocates retrieval attention and budget toward high‑value aspects, thereby retrieving a more informative and balanced set of evidence pages. The final answer is generated by reasoning over the retrieved pages and their correlations. MAB‑DQA outperforms existing methods on four benchmarks, with a 10.38\% average gain in answer accuracy over the strongest baseline, and achieves best retrieval performance. Our contributions are:
\begin{itemize}[leftmargin=0.2cm, itemindent=0cm]
\item We propose MAB‑DQA, a novel multi‑armed bandit-based multi-modality DQA framework that advances multi‑aspect query by dynamically discerning the importance of query aspects, thereby guiding retrieval to prioritize evidence containing critical information.
\item We design a bandit‑guided retrieval strategy that treats each aspect‑aware subquery as an arm and uses preliminary reasoning feedback from vision-language model as rewards, enabling adaptive exploration-exploitation in page retrieval.
\item We conduct extensive experiments on the four open-source benchmarks, demonstrating that MAB‑DQA consistently enhances multi-modality document understanding performance and outperforms existing methods significantly.
\end{itemize}
\section{Related Work}
\label{sec:related_work}

\begin{figure*}[ht]
\centering
\includegraphics[width=1.\textwidth]{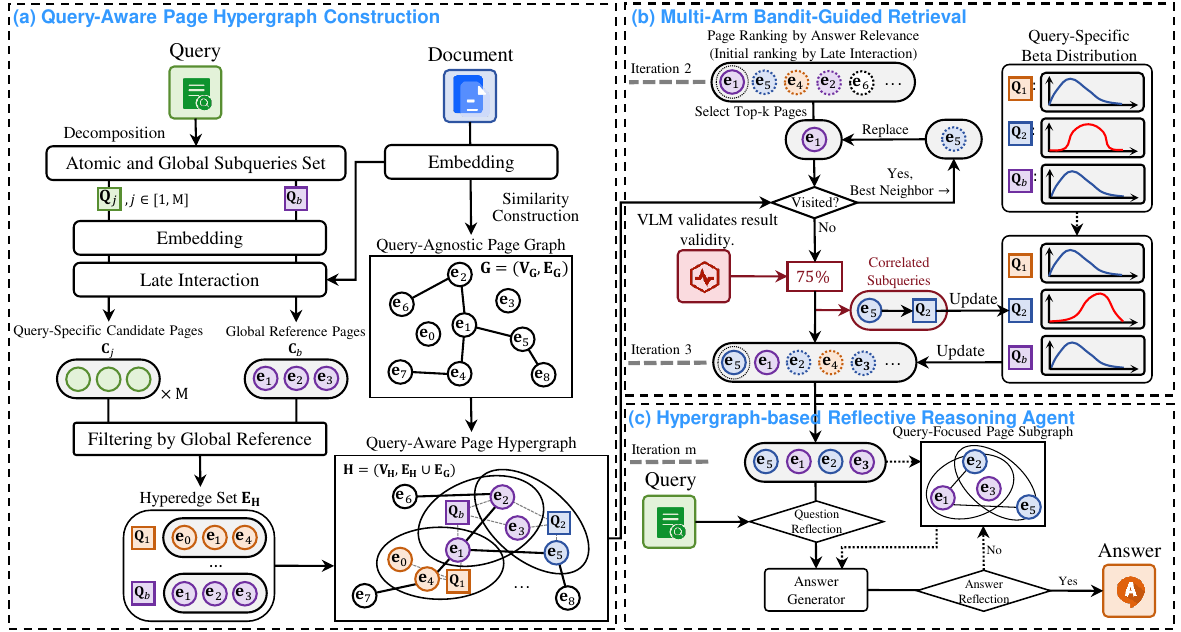}
\caption{The overview of our proposed framework MAB-DQA.
(a) Decompose a query into aspect-aware subqueries and model the relationships using hyperedges to construct a hypergraph;
(b) For the hypergraph's jumping retrieval pages, model each subquery as an arm and use the Bandit-based method to select the next page based on VLM feedback;
(c) Derive the final answer through reflective reasoning.}
\label{fig:frame}
\end{figure*}

\noindent\textbf{Document Question Answering.} DQA serves as a key task for evaluating models' document comprehension capabilities~\citep{acl:chunk:1, acl:qa:5}.
With the advancement of large language models (LLMs) and vision language models (VLMs), research focus has shifted from short-form unimodal to long-form multimodal understanding~\citep{acl:llm:1, shen2026imaggarment, shen2025triplet}.
LLMs such as GPT-4o~\citep{gpt_4o} and Qwen-VL~\citep{qwen} support long document inputs by extending context windows, but suffer from information dilution~\citep{acl:qa:3, acl:qa:6, acl:qa:7, acl:qa:8}.
For this, RAG is a mainstream paradigm that improves quality via external indexes.

\noindent\textbf{Retrieval-Augmented Generation (RAG).} A common strategy in RAG is the "chunk-vectorize" approach, which uses models like BERT~\citep{bert}, Colpali~\citep{colpali}, and ColBERT~\citep{colbert} to generate embeddings, and leverages LLMs for optimization. For instance, MDocAgent~\citep{mdocagent} employs multi-agent collaboration for DQA~\citep{acl:qa:2}. Still, chunk-based RAG fails to represent complex relationship~\citep{tang2019adversarial}.

\noindent\textbf{Graph-based RAG and Query Decomposition.} Graph-based RAG~\citep{graph_rag} has gained attention for enhancing reasoning through knowledge graphs and relational paths~\citep{acl:graph:1}. Simultaneously, research on query decomposition and query rewriting has also progressed~\citep{zhang_retrieve_2023}.
Some studies advance from the perspectives of GraphRAG or query decomposition. For example, Cog-RAG~\citep{cog_rag} constructs hypergraph indexes to improve topic consistency; Self-RAG~\citep{self_rag} trains models to autonomously evaluate retrieval and query quality; RA-DIT~\citep{ra_dit} jointly optimizes the retriever and generator; DRAG~\citep{drag} deconstructs complex query risks. MBA-RAG~\citep{mba-rag} also employs MAB but differs from MAB-DQA, focusing on unimodal contexts, where MAB selects retrieval strategies for cost control. MoloRAG~\citep{molo_rag} is a state-of-the-art multimodal method that enhances DQA by constructing a graph and applying a VLM for graph traversal. It advances retrieval through modeling page relationships, yet operates under a fixed retrieval budget~\citep{fu2026dmap}. A key drawback is uniform attention across queries, ignoring subtle but valuable content~\citep{du2026uncertainty}. However, existing methods are unable to replicate the ability of humans to evaluate aspects of a query while visually reading documents.
\section{Methodology}

We propose a Multi-Armed Bandit–based Document Question Answering framework (MAB-DQA). As illustrated in Fig.~\ref{fig:frame}, the key to our approach is the explicit modeling of aspect‑aware subqueries and the dynamic allocation of retrieval effort among them. To achieve this, we first decompose the query and represent the document as a hypergraph (Sec.~\ref{sec:qdhc}, Fig.~\ref{fig:frame}(a)), where hyperedges represent an aspect-specific candidate set relevant to each subquery. Subsequently, a MAB mechanism treats each subquery as an arm and uses preliminary Vision‑Language Model (VLM) feedback as rewards to guide an exploration-exploitation policy over the hypergraph (Sec.~\ref{sec:bpn}, Fig.~\ref{fig:frame}(b)). Finally, the answer is obtained through multi‑stage verification via a Hypergraph‑based Reflective Reasoning Agent (Sec.~\ref{sec:hrra}, Fig.~\ref{fig:frame}(c)).

\subsection{Preliminary: Late Interaction Retrieval}

Given a query $ q $ and $ N $ document pages $p_i$, their multi-vector embeddings in $ \mathbb{R}^D $ are $ \mathbf{E_q} \in \mathbb{R}^{N_q \times D} $ and $ \mathbf{E_p} \in \mathbb{R}^{N_p \times D} $ per page, with $ N_q $ and $ N_p $ as vector counts. The Late Interaction (LI)~\cite{colbert} operator $ \mathbf{LI}(q,p) $ computes for each query vector $ \mathbf{E_q}^{(k)} $ the maximum dot product with all $ \mathbf{E_p}^{(l)} $, denoted $ \text{max}\langle \cdot | \cdot \rangle $, and sums these:
\begin{align}\label{eq:li}
\mathbf{LI}(q,p_i) = \sum^{N_q}_{k=1} \max^{N_p}_{l=1} \langle \mathbf{E_q}^{(k)} | \mathbf{E_{p_i}}^{(l)} \rangle.
\end{align}

A limitation of this formulation is that it assigns equal weight to every query vector $\mathbf{E}_q^{(k)}$, which may not reflect the varying importance of different semantic aspects in the query.

\subsection{Query-Aware Page Hypergraph}
\label{sec:qdhc}

To capture the multi‑aspect nature of complex queries, we first build a Query-Agnostic Page Graph $ \mathbf{G}(\mathbf{V}_\mathbf{G},\mathbf{E}_\mathbf{G}) $ to represent the relationships between pages.
Each node $ p_i \in \mathbf{V}_\mathbf{G} $ corresponds to a page.
An edge $ \mathbf{E}_\mathbf{G} $ is added between nodes $ \{ p_{i},p_{j} \}, i \neq j $ if the similarity between the two pages exceeds the threshold $ \theta_{\mathbf{G}} $, expressed as:
\begin{align}
\mathbf{E}_\mathbf{G} = \{ \{ p_{i},p_{j} \} \mid \text{sim}\langle \mathbf{E}_{p_{i}}, \mathbf{E}_{p_{j}} \rangle \geq \theta_{\mathbf{G}} \},
\end{align}
where $ \text{sim}\langle \cdot, \cdot \rangle $ denotes the inner product.
Based on the graph $ \mathbf{G} $, a VLM rewrites the original query $ q $ and decomposes it into a set of subqueries $ \mathcal{E}_q = \{q_1, q_2, ..., q_M\} $, as detailed in Appendix~\ref{sec:appendix_decompose}.
We then define the Atom‑Integral Subqueries Set as:
\begin{align}
\mathbf{Q} = \{ \{ \hat{q} \} \mid \hat{q} \in \mathcal{E}_{q} \}\cup \{\mathcal{E}_{q}\},
\end{align}
which includes both fine‑grained (atomic) subqueries and the global query $\mathcal{E}_q$ (denoted as $\mathcal{E}_{M+1}$).
We select the top‑$\theta_{\mathbf{H}}$ pages with the highest $\mathbf{LI}(\mathbf{Q}_j,p_i)$ scores to form the Query‑Specific Candidate Pages set $ \textbf{C}_j $.
We then select pages not in the global reference candidate pages set $ \textbf{C}_b, \ b=M+1 $, or those in both $ \textbf{C}_j $ and $ \textbf{C}_b$, with better ranking under $ \textbf{Q}_j $, to construct a hyperedge:
\begin{align}
\hat{\mathbf{E}}_j = \{p \in \mathbf{C}_j \mid p \notin \mathbf{C}_b \vee \text{rank}(\mathbf{LI}(\mathbf{Q}_j,p)) \nonumber \\
\leq \text{rank}(\mathbf{LI}(\mathbf{Q}_b,p)) \},
\end{align}
where $\text{rank}(\mathbf{LI}(\cdot, p))$ denotes the descending‑order position of page $p$ under the corresponding query based on its $\mathbf{LI}$ score.
Finally, the Query-Aware Page Hypergraph is defined as:
\begin{align}
\mathbf{H}(\mathbf{V}_\mathbf{H},\mathbf{E}_\mathbf{H} \cup \mathbf{E}_\mathbf{G}) = (\mathbf{V}_\mathbf{G},\{\hat{\mathbf{E}}_j\}^{M+1}_{j=1} \cup \mathbf{E}_\mathbf{G}).
\end{align}

\subsection{Multi-Arm Bandit-Guided Retrieval}
\label{sec:bpn}

We frame the retrieval process over $\mathbf{H}$ as a combinatorial multi‑armed bandit problem. Each subquery $\mathbf{Q}_j$ is treated as an arm, and the goal is to sequentially decide which arms (subqueries) to "pull" in order to retrieve the most informative pages.

\noindent\textbf{Reward Model and Thompson Sampling.} When the VLM inspects a retrieved page $p_i$, it produces a relevance score $s^{\text{vlm}}_i \in [0,1]$ indicating whether the page contains useful evidence for the query~\cite{molo_rag}, as detailed in Appendix \ref{sec:appendix_retrieval_evidence}. This score serves as the reward signal for the bandit.

Each arm $\mathbf{Q}_j$ maintains a Beta distribution $\text{Beta}(\alpha_j, \beta_j)$ to model its reward probability. Initially, $\alpha_j = \beta_j = 1$ (uniform prior). The probability density function is:
\begin{align}
f(x;\alpha_j, \beta_j) &= \frac{1}{\text{Beta}(\alpha_j, \beta_j)}x^{\alpha_j - 1}(1 - x)^{\beta_j - 1} \\
\text{Beta}(\alpha_j, \beta_j) &= \int^{1}_{0}t^{\alpha_j - 1}(1 - t)^{\beta_j - 1}dt \nonumber\\
&= \frac{\Gamma(\alpha_j)\Gamma(\beta_j)}{\Gamma(\alpha_j + \beta_j)},
\end{align}
where $ \Gamma(\cdot) $ is the Gamma function, and $x \in [0,1]$.
The initial parameters are set as $ \alpha_j = 1 $ and $ \beta_j = 1 $. We employ Thompson sampling~\citep{ts:1,ts:2} to balance exploration and exploitation. In each retrieval step, a sample is drawn from each arm’s Beta distribution, and the arm with the highest sample value is selected to guide the subsequent page retrieval.

\noindent\textbf{Scoring and Node Expansion.} During a jumping retrieval of $\mathbf{H}$, each page node $p_i$ receives a composite score for sorting:
\begin{align}
\text{score}(p_i) &= (1 - \alpha) \max_{j \in [1,M+1]}\mathbf{LI}(\mathbf{Q}_j,p_i) \nonumber \\
&+ \alpha s^{\text{vlm}}_i + \beta [(1-\lambda) h_i + \lambda \bar{s}^{\text{cb}}_i],
\end{align}
where $ h_i $ is the degree of page $ p_i $ in $ \mathbf{H} $.
$ \bar{s}^{\text{cb}}_i $ denotes the Thompson Sampling confidence score of the associated subqueries, calculated as:
\begin{align}
\bar{s}^{\text{cb}}_i = \frac{1}{|\hat{\mathbf{Q}}_i|} \sum_{\mathbf{Q}_j \in \hat{\mathbf{Q}}_i} \mathbb{E}[\text{Beta}(\alpha_j,\beta_j)],
\end{align}
where $ \hat{\mathbf{Q}}_i $ denotes all subqueries linked to page $ p_i $ via hyperedges in $ \mathbf{H} $.
Generally, $ h_i $ emphasizes the page’s own contribution, whereas $ \bar{s}^{\text{cb}}_i $ focuses more on the contribution of subqueries.
A larger $ \alpha \in [0,1] $ gives greater weight to the VLM evaluation results.
The $ \beta $ is used to adjust the proportion between hyperparameters.
The $ \lambda \in [0,1] $ balances the page degree and arm confidence;
Larger $ \lambda $ values favor retrieval toward the highest overall subquery combinations expectation across a subquery rather than a single page.

The top‑$k$ nodes with the highest scores are expanded in each round. After the VLM evaluates a retrieved page $p_i$ and returns $s^{\text{vlm}}_i$, all arms $\mathbf{Q}_j$ associated with $p_i$ update their parameters:
\begin{align}
\forall \mathbf{Q}_j \in \hat{\mathbf{Q}}_i,  (\alpha_j, \beta_j) \leftarrow (\alpha_j + s^{\text{vlm}}_i, \beta_j + 1 - s^{\text{vlm}}_i).
\end{align}

This update increases $\alpha_j$ if the page is relevant (high VLM score) and increases $\beta_j$ otherwise, thereby refining the reward estimate for each aspect‑specific subquery.

\subsection{Hypergraph-based Reflective Reasoning Agent}
\label{sec:hrra}

After the MAB-guided retrieval obtains a set of relevant pages from the hypergraph $\mathbf{H}$, the Hypergraph‑based Reflective Reasoning Agent (HRRA) synthesizes the final answer through a multi‑stage question and answer verification process. Employing an "initial response-verification-optimization" pipeline, HRRA first generates an initial answer using the retrieved evidence. If inconsistencies or evidence gaps are detected, the agent re‑enters the hypergraph and constructs the Query-Focused Page Subgraph in a reflective loop. 
\section{Experiment}

\subsection{Implementation Details}

The main experiments employ the QWen-2.5VL-7B-Instruct model~\citep{qwen} for both retrieval-augmented generation and question answering.
Additional vision-language models are also included in the ablation studies.
For generating embeddings, the ColPali~\citep{colpali} model is selected as a vision-language embedding model specifically optimized for documents.
All embedding generation, information retrieval, and question answering processes are conducted on a system equipped with four NVIDIA V100 GPUs.

\begin{table}[ht]
\caption{Statistics of Datasets. "Issue" refers to the number of samples in the labeled dataset that the Colpali model may be ignoring regarding key query conditions.}
\centering
\small
\begin{tabular}{l|ccc}
\toprule
\textbf{Dataset} & \textbf{Document} & \textbf{Question} & \textbf{Issue} \\
\midrule
MMLongBench &
134 & 1073 & \textbf{212}(19.8\%) \\

LongDocURL &
396 & 2325 & \textbf{647}(27.8\%) \\

PaperTab &
307 & 393 & - \\

FetaTab &
871 & 1016 & - \\

\bottomrule
\end{tabular}
\label{tab:dataset}
\end{table}

\begin{table*}[!t]
\caption{Comparison of our method with DQA approaches. Pure VLM-based, Multi-Agent-based, and RAG-based methods are evaluated using accuracy (\%).}
\centering
\small
\begin{tabular}{lcccccc}
\toprule
\textbf{Model} & \textbf{RAG Method} 
& \textbf{MMLongBench} & \textbf{LongDocURL} & \textbf{FetaTab} & \textbf{PaperTab} 
& \textbf{Avg} \\
\midrule
Qwen-2.5-VL-7B
& Direct
& 0.204 & 0.398 & 0.350 & 0.112 & 0.266 \\

LLaVA-1.6-7B
& Direct
& 0.176 & 0.110 & 0.301 & 0.102 & 0.172 \\

\midrule
MDocAgent 
& ColPali + ColBert 
& 0.315 & 0.527 & 0.598 & 0.227 & 0.417 \\

\midrule
M3DocRAG
& ColPali (Top-4) 
& 0.296 & 0.503 & 0.537 & 0.152 & 0.372 \\

MoloRAG
& ColPali (Top-4)  
& 0.371 & 0.536 & 0.554 & 0.157 & 0.405 \\

MoloRAG+
& MoloRAG+ (Top-4)  
& 0.372 & 0.528 & 0.600 & 0.195 & 0.424 \\

\rowcolor[gray]{0.9}
\textbf{MAB-DQA (Ours)}
& \textbf{ColPali (Top-4)} 
& \textbf{0.399} & \textbf{0.564} & \textbf{0.638} & \textbf{0.269} & \textbf{0.468} \\
\midrule
\multicolumn{2}{c}{\textbf{Improvement over second-best (\%)}} 
& \textbf{+7.25\%} & \textbf{+5.22\%} & \textbf{+6.33\%} & \textbf{+18.50\%} & \textbf{+10.38\%} \\
\bottomrule
\end{tabular}
\label{tab:main_qa_results}
\end{table*}

\begin{table*}[!t]
\caption{Retrieval Performance Comparison on MMLongBench and LongDocURL Benchmarks under Top-K ($K=1,3,5$) Settings. All values are in \%. The best results are in bold.}
\centering
\small
\begin{tabular}{cc|cccc|cccc}
\toprule
\multirow{2}{*}{\textbf{Top-K}} & \multirow{2}{*}{\textbf{Method}} & \multicolumn{4}{c|}{\textbf{MMLongBench}} & \multicolumn{4}{c}{\textbf{LongDocURL}} \\
 & & \textbf{Recall} & \textbf{Precision} & \textbf{NDCG} & \textbf{MRR} & \textbf{Recall} & \textbf{Precision} & \textbf{NDCG} & \textbf{MRR} \\
 
\midrule

\multirow{6}{*}{1}
& M3DocRAG
& 45.31 & 56.80 & 56.80 & 56.80
& 46.82 & 64.51 & 64.45 & 64.51 \\

& MDocAgent (ColBert)
& 30.65 & 40.50 & 40.50 & 40.50
& 40.72 & 56.31 & 56.31 & 56.31 \\

& MDocAgent (Colpali)
& 46.61 & 59.86 & 59.86 & 59.86
& 46.90 & 64.18 & 64.18 & 64.18 \\

& MoLoRAG
& 48.93 & 64.11 & 64.11 & 64.11
& 49.59 & 67.84 & 67.84 & 67.84 \\

& MoLoRAG+
& 50.30 & 64.82 & 64.82 & 64.82
& 48.70 & 66.90 & 66.90 & 66.90 \\

\rowcolor[gray]{0.9}\cellcolor{white}
& MAB-DQA (Ours)
& \textbf{50.97} & \textbf{66.35} & \textbf{66.35} & \textbf{66.35}
& \textbf{50.60} & \textbf{69.95} & \textbf{69.95} & \textbf{69.95} \\

\midrule

\multirow{6}{*}{3}
& M3DocRAG
& 64.69 & 32.74 & 38.04 & 65.47
& 66.98 & 33.53 & 38.79 & 72.51 \\

& MDocAgent (ColBert)
& 45.70 & 21.92 & 30.84 & 47.64
& 56.70 & 28.35 & 39.84 & 63.44 \\

& MDocAgent (Colpali)
& 65.90 & 32.47 & 38.42 & 67.73
& 68.14 & 34.45 & 40.91 & 72.92 \\

& MoLoRAG
& 67.40 & 33.18 & 39.70 & 70.66
& 69.58 & 35.27 & 42.30 & 75.60 \\

& MoLoRAG+
& 66.95 & 33.10 & 39.97 & 71.02
& 69.42 & 35.18 & 42.07 & 74.89 \\

\rowcolor[gray]{0.9}\cellcolor{white}
& MAB-DQA (Ours)
& \textbf{69.53} & \textbf{34.32} & \textbf{41.05} & \textbf{72.94}
& \textbf{70.46} & \textbf{35.73} & \textbf{43.07} & \textbf{77.78} \\

\midrule

\multirow{6}{*}{5}
& M3DocRAG
& 72.43 & 22.67 & 30.06 & 66.92
& 74.54 & 23.52 & 32.07 & 73.99 \\

& MDocAgent (ColBert)
& 53.15 & 16.06 & 26.43 & 49.27
& 64.53 & 20.09 & 30.60 & 64.88 \\

& MDocAgent (Colpali)
& 73.07 & 23.07 & 30.25 & 69.04
& 75.49 & 23.87 & 31.81 & 74.30 \\

& MoLoRAG
& 71.42 & 21.96 & 30.13 & 71.27
& 73.86 & 23.20 & 32.01 & 76.18 \\

& MoLoRAG+
& 70.32 & 21.42 & 29.99 & 71.58
& 73.69 & 23.15 & 31.83 & 75.56 \\

\rowcolor[gray]{0.9}\cellcolor{white}
& MAB-DQA (Ours)
& \textbf{75.86} & \textbf{24.13} & \textbf{32.14} & \textbf{73.85}
& \textbf{77.02} & \textbf{24.30} & \textbf{33.13} & \textbf{78.73} \\

\bottomrule
\end{tabular}
\label{tab:main_rag_results}
\end{table*}

\noindent\textbf{Datasets.} To validate the effectiveness of the proposed method, experiments were conducted on four benchmark datasets, which are briefly described below.
MMLongBench~\citep{mmlongbench} comprehensively evaluates model document understanding capabilities. Its questions often require cross-page reasoning (33.7\%) and include approximately 20.6\% unanswerable questions, specifically designed to detect "hallucination" tendencies in DQA systems.
LongDocURL~\citep{longdocurl} is a multimodal dataset focused on long document processing. It contains a large number of cross-modal questions to assess model performance in long-text contexts.
PaperTab~\citep{ptab} is a specialized dataset for scientific paper comprehension, with particular emphasis on interpreting document structure and tabular data.
FetaTab~\citep{fetaqa} is a Wikipedia-based question answering dataset that includes rich tabular and chart information.
Key statistics for all datasets are listed in Table~\ref{tab:dataset}.
Furthermore, we perform a dedicated analysis based on the labeled data.
The "Issue" column in Table~\ref{tab:dataset} shows the proportion of retrieval errors in the Colpali model caused by ignoring key query constraints.
The judging rule is: an error is counted if the retrieval performance for a subquery $\mathbf{Q}_j$ is better than that for the original query $q$.

\noindent\textbf{Baselines.} We selected three types of frameworks as baselines:
A pure VLM-based DQA framework;
A multimodal DQA framework based on multi-RAG and multi-agent, represented by MDocAgent~\citep{mdocagent};
A multimodal RAG-based DQA system.
In the pure VLM approach, documents are directly provided as context to the VLM for question answering.
In the multimodal RAG approach, M3DocRAG~\citep{m3docrag}, and the MoloRAG~\citep{molo_rag}, which also employs graph structures and VLM evaluation, were selected.
For fair comparison, MoloRAG uses a 7B model instead of the 3B model.
The MoloRAG+ model, compared to MoloRAG, employs a fine-tuned model for retrieval.

\noindent\textbf{Metrics.} For the DQA evaluation metrics, the experiments are the same as those used in MDocAgent, LongDocURL, and MMLongBench, employing GPT‑4o~\citep{gpt_4o} to assess the outputs.
Given a question and its reference answer, GPT‑4o compares the DQA system’s output and returns a Boolean value indicating whether the answer is correct and complete.
For the RAG evaluation metrics, the experimental evaluation metrics align with those in MoloRAG, including Recall, Precision, Normalized Discounted Cumulative Gain (NDCG), and Mean Reciprocal Rank (MRR).

\subsection{Main Results}

The comparative results between our method and various baselines are presented in Table~\ref{tab:main_qa_results} and Table~\ref{tab:main_rag_results}.
Our approach consistently outperforms all baseline methods across the four evaluated datasets in terms of question answering accuracy, achieving an average improvement of 10.38\% over the strongest baseline.
Notably, the performance gain is especially pronounced on the PaperTab dataset (+18.50\%), which emphasizes the understanding of document structure and tables.

Moreover, as shown in Table~\ref{tab:main_rag_results}, our method also establishes a new state-of-the-art in retrieval performance on both the MMLongBench and LongDocURL benchmarks, surpassing existing methods across all Top‑K settings and all retrieval metrics (Recall, Precision, NDCG, and MRR). Our proposed method outperforms the baseline (Colpali) by an average of 6.55\% across all metrics.
These results demonstrate the effectiveness of MAB-DQA in enhancing both the precision of retrieval and the accuracy of answer generation in DQA tasks.

\subsection{Ablation Studies}
\label{sec:ablation}

\begin{table*}[!t]
\caption{Ablation Study on the Proposed MAB-DQA Modules. Additional ablation in the Appendix.}
\centering
\small
\begin{tabular}{l|c@{ }c@{ }c|cccc|c}
\toprule
\textbf{Model Variant} &
    \textbf{QD} & \textbf{MABR} & \textbf{HRRA} &
\textbf{MMLongBench} & \textbf{LongDocURL} & \textbf{FetaTab} & \textbf{PaperTab} &
\textbf{Avg. Imp.} \\
\midrule

Colpali (Baseline) & 
$\times$ & $\times$ & $\times$ &
0.296 & 0.554 & 0.537 & 0.152 &
0.0\% \\

w/ MABR &
$\circ$ & $\circ$ & $\times$ &
0.388 & 0.543 & 0.609 & 0.226 &
22.8\% \\

w/ HRRA &
$\circ$ & $\times$ & $\circ$ &
0.395 & 0.561 & 0.624 & 0.236 &
26.5\% \\

MAB-DQA (Ours) &
$\circ$ & $\circ$ & $\circ$ &
\textbf{0.399} & \textbf{0.564} & \textbf{0.638} & \textbf{0.269} &
33.1\% \\

\bottomrule
\end{tabular}
\label{tab:ablation}
\end{table*}

This section evaluates the contributions of two core modules: Multi-Arm Bandit-Guided Retrieval (MABR) and Hypergraph‑based Reflective Reasoning Agent (HRRA) to the framework performance through ablation experiments. As in Table~\ref{tab:ablation}, when only basic retrieval is used (Colpali), the performance is the weakest. This indicates that the lack of differentiation among condition importance leads to retrieval interference from secondary information, severely limiting the accuracy of evidence localization and answer generation.
When using MABR (w/ MABR), performance improves (an average gain of 22.8\%), yet remains significantly lower than the full model. This demonstrates that the absence of a dynamic path selection mechanism reduces retrieval precision and robustness.
When query decomposition (QD) guided retrieval and HRRA is retained, but MAB is removed (w/ HRRA), the model achieves an average improvement of 26.5\%, yet still underperforms on complex questions requiring aspect reasoning. This suggests that reflective reasoning and multi‑stage verification are essential for information validation and error correction. The full model (Ours) achieves the best performance across all datasets, with an average improvement of 33.1\%; the gain from HRRA is most pronounced on complex datasets. MABR and HRRA progressively build upon and reinforce each other: MABR enables adaptive retrieval focusing on key aspects, and HRRA further integrates and validates information through reflective reasoning.

\begin{table*}[!t]
\caption{Evaluation on MMLongBench and LongDocURL across VLM Backbones and Methods. Retrieval performance was evaluated based on Top-3 settings.}
\centering
\small
\begin{tabular}{lc|cccc|cccc}
\toprule
\multirow{2}{*}{\textbf{Backbone}} & \multirow{2}{*}{\textbf{Method}} & \multicolumn{4}{c|}{\textbf{MMLongBench}} & \multicolumn{4}{c}{\textbf{LongDocURL}} \\
 & & \textbf{Recall} & \textbf{Precision} & \textbf{NDCG} & \textbf{MRR} & \textbf{Recall} & \textbf{Precision} & \textbf{NDCG} & \textbf{MRR} \\
\midrule
Qwen2.5-VL-7B & 
\multirow{4}{*}{MoloRAG}
& 67.40 & 33.18 & 39.70 & 70.66 
& 69.58 & 35.27 & 42.30 & 75.60 \\

LLaVa-13B &
& 65.46 & 32.23 & 38.05 & 67.00
& - & - & - & - \\

Qwen3-30B-A3B &
& 68.41 & 34.08 & 40.13 & 70.78
& 69.80 & 36.02 & 42.47 & 75.77 \\

Qwen3-32B &
& 72.67 & 35.46 & 42.23 & 73.34  
& 70.57 & 35.93 & 43.40 & 77.95 \\

\midrule

Qwen2.5-VL-7B &
\multirow{4}{*}{Ours}
& 69.53 & 34.32 & 41.05 & 72.94
& 70.46 & 35.73 & 43.07 & 77.78 \\

LLaVa-13B &
& 66.24 & 32.37 & 38.21 & 67.12 
& - & - & - & - \\

Qwen3-30B-A3B &
& 72.12 & 35.60 & 42.15 & 74.23 
& 70.97 & 35.97 & 43.09 & 77.52 \\

Qwen3-32B &
& \textbf{73.05} & \textbf{36.25} & \textbf{43.19} & \textbf{76.09}
& \textbf{73.65} & \textbf{36.11} & \textbf{43.91} & \textbf{80.45}  \\

\bottomrule
\end{tabular}
\label{tab:diff_model}
\end{table*}

Meanwhile, we conducted an ablation study on the performance of different VLMs under the MAB-DQA framework, as shown in Table~\ref{tab:diff_model}.
In the table, we used the MoloRAG model as a reference and considered the top-3 retrieval results as the evaluation scope.
This experiment was designed to evaluate whether "measuring the importance of aspects" is essential for different VLM backbones in DQA tasks. We selected four models: Qwen2.5-7B~\citep{qwen}, Llava-13B~\citep{llava}, Qwen3-30B (a MoE model), and Qwen3-32B.

\subsection{Sensitivity Analysis}
\label{sec:sensitivity}

\begin{figure}[ht]
\centering
\includegraphics[width=1.\linewidth]{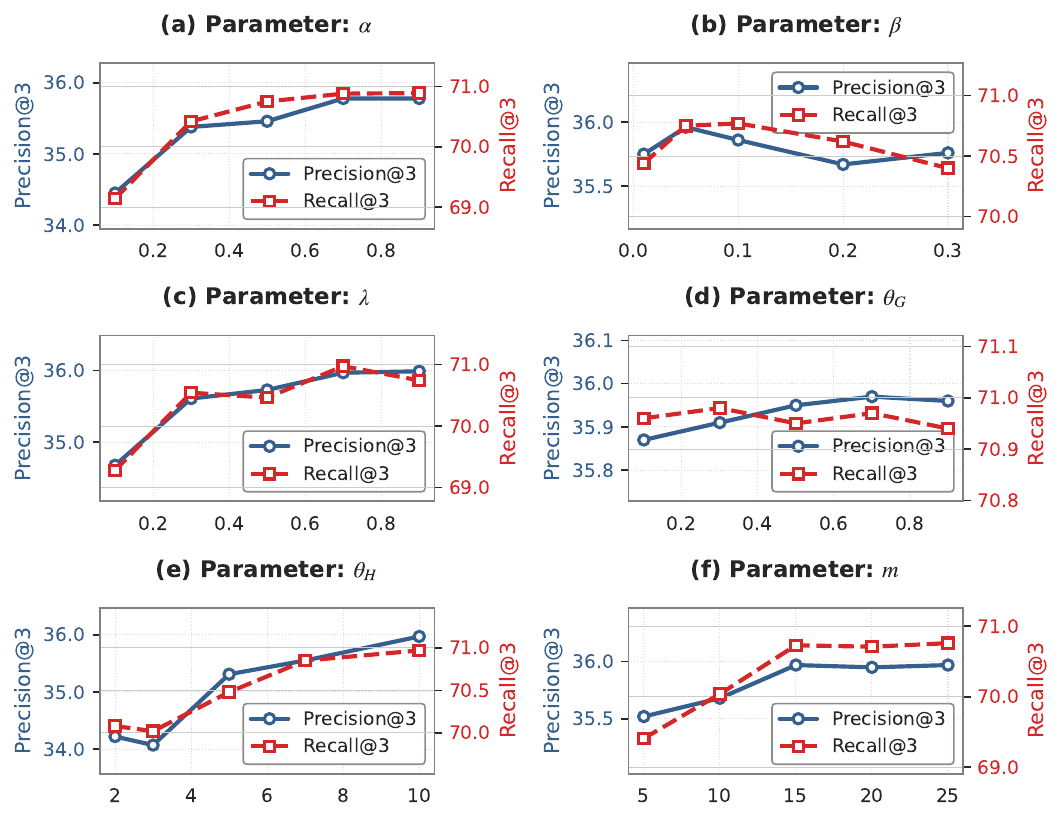}
\caption{Sensitivity Analysis of Key Hyperparameters of the MAB-DQA Framework under Top-3 Retrieval on LongDocURL. The blue dashed line represents Precision. The red line represents Recall.}
\label{fig:sensitivity}
\end{figure}

\begin{figure}[ht]
\centering
\includegraphics[width=1\linewidth]{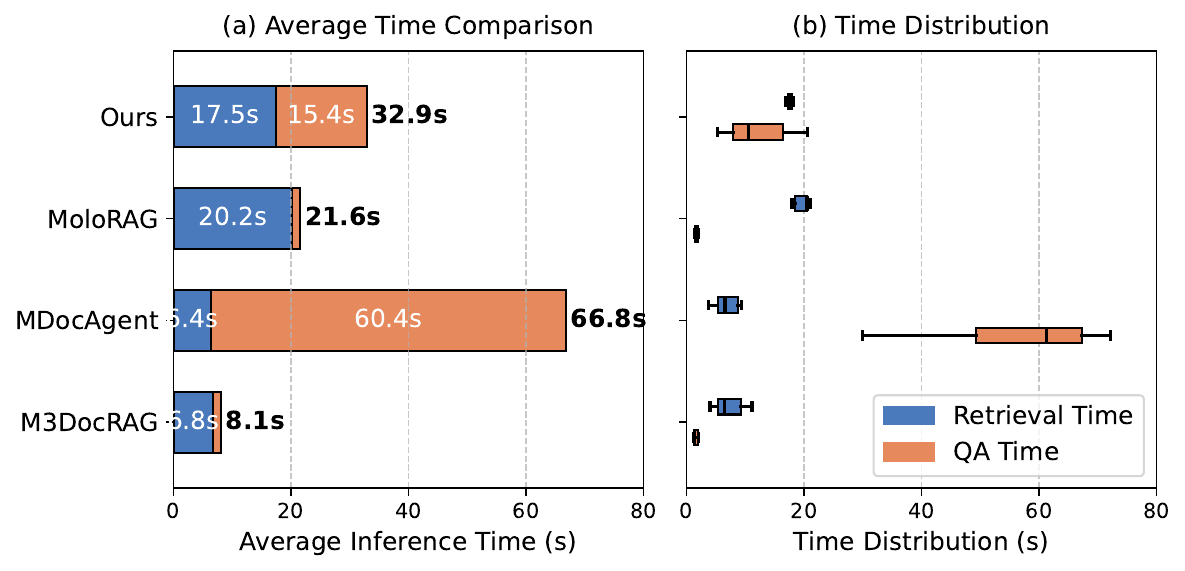}
\caption{Comparison of Average Inference Time Across Different Models on 4 NVIDIA V100 GPUs, with retrieval time (Top-10) in blue and QA time (Top-4) in orange.}
\label{fig:inference}
\end{figure}

\begin{figure*}[!t]
\centering
\includegraphics[width=1\linewidth]{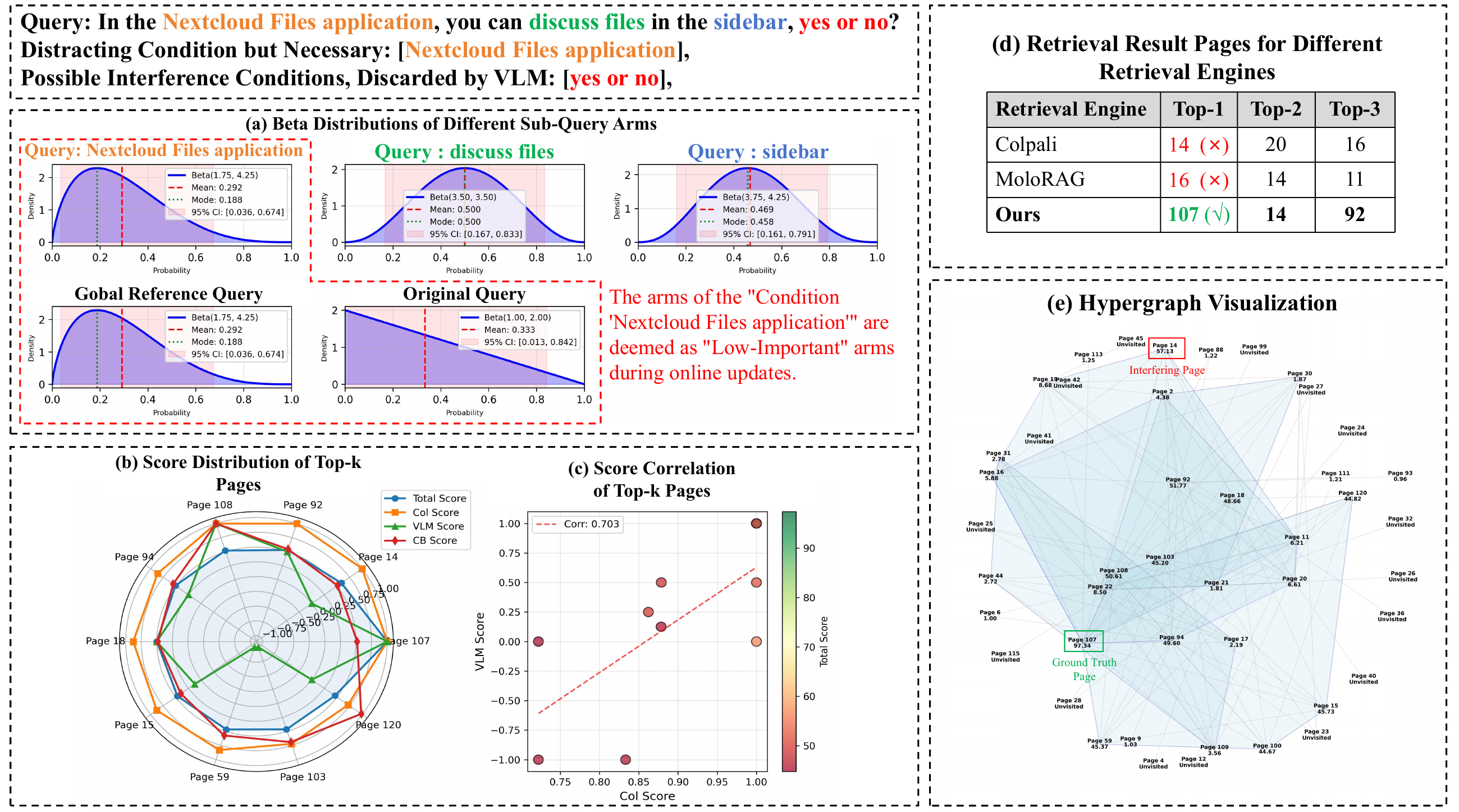}
\caption{Qualitative Analysis of the MAB-DQA Framework.}
\label{fig:case}
\end{figure*}

To evaluate the robustness of the MAB-DQA framework to key hyperparameters, we conducted a systematic sensitivity analysis on the LongDocURL dataset (Fig.~\ref{fig:sensitivity}). Using a controlled variable approach, we adjusted one target parameter at a time and observed changes in Recall and Precision under the Top-3 retrieval setting, while keeping all other parameters fixed.
The results indicate that: (1) Parameter $\alpha$ has a significant positive effect on performance (Fig.~\ref{fig:sensitivity}a), with higher values better leveraging the visual-language model (VLM) to extract effective semantics; (2) Parameter $\beta$ has a relatively minor impact on performance (Fig.~\ref{fig:sensitivity}b); (3) Parameter $\lambda$ reflects the model’s focus on conditional importance, and increasing it clearly improves performance (Fig.~\ref{fig:sensitivity}c). Additionally, we examined the effects of the edge connection threshold $\theta_{\mathbf{G}}$, hyperedge capacity $\theta_{\mathbf{H}}$, and retrieval iteration number $m$.
Based on the analysis, the hyperparameters are set as follows for subsequent experiments: $\alpha=0.8$, $\beta=0.1$, $\lambda=0.75$, $\theta_{\mathbf{G}}=0.8$, $\theta_{\mathbf{H}}=10$, and $m=20$, ensuring stable and reliable performance across different configurations. Appendix~\ref{sec:appendix_experiment} contains more ablation and sensitivity experiments.

\subsection{Mechanism Analysis}
\label{sec:mechanism}

We provide the complete algorithm table for MAB-DQA in Appendix~\ref{sec:appendix_algorithm}. During the retrieval phase, the time complexity of the MABR algorithm is $O(mT_{\text{VLM}})$, where $T_{\text{VLM}}$ denotes the time cost of VLM evaluation. We also tested the DQA efficiency of different frameworks under the same hardware environment, as shown in Fig.~\ref{fig:inference}.

Qualitative cases of the MAB-DQA are provided to analyze the underlying mechanism of the model.
In Fig.~\ref{fig:case}, we selected a representative multi-aspect query.
For this query, the MAB-DQA framework correctly retrieved the evidence page (Page 107), while the other two baseline methods failed to do so.
In this Fig.~\ref{fig:case}(a), there exists a low-importance condition, "Nextcloud Files application."
The evidence lies in the fact that queries containing this condition (indicated by the red dashed section in Fig.~\ref{fig:case}(a)) are all assigned low rewards by the VLM (the Beta distribution tends towards 0.0).
After correctly evaluating the importance of conditions, MAB-DQA successfully retrieved the evidence on the query-aware page hypergraph (Fig.~\ref{fig:case}(e)).
\section{Conclusion}
\label{sec:conclusion}

This paper addresses a key challenge in multi-aspect DQA, where the retrieval process is often dominated by less important query aspects, leading to the omission of critical evidence. To tackle this, we propose MAB-DQA, a Multi-Armed Bandit-based DQA framework that explicitly models and dynamically allocates attention to the varying importance of implicit aspects within a query. By decomposing the query into aspect-aware subqueries and treating each as an arm in a bandit setup, MAB-DQA uses preliminary reasoning signals to estimate aspect utility and dynamically redistributes retrieval budget toward high-value aspects. 

Extensive experiments on four benchmarks demonstrate that MAB-DQA significantly enhances document understanding performance, achieving an average improvement of 10.38\% in answer accuracy over the strongest baseline. Moreover, MAB-DQA establishes new state-of-the-art retrieval performance, outperforming existing methods across all Top-K settings ($K=1,3,5$) on MMLongBench and LongDocURL benchmarks, as shown in the comprehensive evaluation table.

\section*{Acknowledgements}


This work is supported by Noncommunicable Chronic Diseases-National Science and Technology Major Project (Grant No.~2024ZD0524603) and the Singapore Ministry of Education (MOE) Academic Research Fund (AcRF) Tier 1 grant.

\section*{Limitations}

While the proposed MAB-DQA framework demonstrates significant improvements in multi-aspect Document Question Answering, several limitations remain to be addressed in future work.

\noindent\textbf{Dependence on Visual-Language Model Performance.}
The framework heavily relies on the capability of the underlying VLM for both query decomposition and evidence evaluation. If the VLM performs poorly in specific domains, such as technical, legal, or medical documents, or under low-resource scenarios, the retrieval and reasoning performance may degrade accordingly. 

\noindent\textbf{Scalability with Document Length and Complexity.}
The MAB-DQA method we proposed is applied in our study to long documents (ranging from 40 to 500 pages). If a query requires retrieving a larger number of pages (e.g., as many as over 100 pages of evidence, though this is uncommon), it may be necessary to adjust hyperparameters or perform optimization.

\noindent\textbf{Limitations of Hyperparameter Balancing.} Our method relies on several hyperparameters ($\alpha$, $\beta$, and $\lambda$) to balance different scoring components. Although we selected values via grid search, they may not generalize optimally to all document/query types. Our experiments show that adjustments to these hyperparameters lead to consistent performance fluctuations across multiple datasets. In future work, we plan to develop a version of MAB-DQA that incorporates Bayesian optimization for hyperparameter selection, thereby enhancing its adaptability to diverse DQA scenarios.

\noindent\textbf{Restriction to Thompson Sampling.} Our study exclusively employs Thompson Sampling (TS) as the core bandit algorithm, driven by its principled Bayesian approach, which aligns naturally with the probabilistic reward signals from the VLM. The Bernoulli-like feedback (relevant/irrelevant) is well-modeled by the Beta-Bernoulli conjugate prior, facilitating efficient online updates. However, this focus precludes a comparative analysis against other bandit strategies, such as Upper Confidence Bound (UCB) or Epsilon-Greedy. UCB offers stronger theoretical regret bounds and deterministic action selection, which might provide more stable retrieval paths. Epsilon-Greedy, while simpler, could be more effective in highly non-stationary environments where query aspect importance shifts rapidly. The superiority of TS in our specific combinatorial bandit setting has been empirically observed. Future work should include a comprehensive study to explore adaptive mechanisms that dynamically select the most suitable bandit strategy based on query characteristics.

\section*{Ethical Statement}

In this work, we utilize models and datasets sourced from open‑source platforms. All models and datasets are licensed under the Creative Commons Attribution 4.0 International License (CC BY 4.0). Their use fully complies with the corresponding license terms and is strictly limited to academic and research purposes. No private, sensitive, or personally identifiable information was employed in this study. Our research adheres to the ACL Code of Ethics and follows the ACL ethics guidelines, ensuring integrity, transparency, and reproducibility throughout the work.


\appendix
\section{Details of Evaluation Metrics}

This paper employs two categories of evaluation metrics: \textbf{Document-based Question Answering accuracy metrics} and \textbf{Retrieval-Augmented Generation retrieval metrics}. The following details the definition, calculation, and application of each metric in the experiments.

\subsection{Document-based Question Answering Accuracy Metrics}

While the automated evaluation paradigm described herein is tailored for DQA, the principle of leveraging powerful pre-trained models for assessment shares conceptual ground with evaluation challenges in cross-modality generation tasks~\citep{xiang2025cfdm}. The DQA metrics are used to evaluate the correctness of the model-generated answers. The evaluation is specifically based on automated assessment using GPT-4o. The process is as follows: Given a question and its reference answer, GPT-4o compares the system output with the reference answer and judges whether the answer is "correct and complete." The evaluation results in a binary score (0 or 1). Score 1: The output answer is consistent with the reference answer in terms of facts and logic (including handling "unanswerable" questions; if the reference answer is "Not answerable" and the model output is the same, it receives a score of 1). Score 0: The answer is incorrect or lacks key information. The final DQA accuracy is defined as the proportion of correctly answered questions to the total number of questions, as shown in Table~\ref{tab:main_qa_results} of the main text. This metric emphasizes the precision of the answer rather than partial correctness.

\subsection{Retrieval-Augmented Generation Retrieval Metrics}

The retrieval metrics are used to evaluate the performance of the hypergraph retrieval module. They include Recall, Precision, Normalized Discounted Cumulative Gain (NDCG), and Mean Reciprocal Rank (MRR). These metrics are computed under Top-K ($K=1, 3, 5$) settings. Assuming the retrieval engine obtains a predicted page sequence $\hat{p}^{\text{pred}}$ and a ground-truth evidence page sequence $\hat{p}^{\text{gt}}$, the metrics are defined as follows:

\noindent \textbf{Recall}: Measures the proportion of retrieved relevant pages to all relevant pages.
\begin{align}
\operatorname{Recall@K} = \frac{|\hat{p}^{\text{pred}}@K \cap \hat{p}^{\text{gt}}|}{|\hat{p}^{\text{gt}}|},
\end{align}
where $\hat{p}^{\text{pred}}@K$ denotes the top-$K$ retrieved pages, and $|\cdot|$ denotes the cardinality of a set.

\noindent \textbf{Precision}: Measures the proportion of relevant pages among the retrieved results.
\begin{align}
\operatorname{Precision@K} = \frac{|\hat{p}^{\text{pred}}@K \cap \hat{p}^{\text{gt}}|}{K}.
\end{align}

\noindent \textbf{Normalized Discounted Cumulative Gain (NDCG)}: Evaluates the quality of the retrieval ranking by considering positional weighting for relevant pages. The relevance score is based on the LI score.
\begin{align}
\operatorname{DCG@K} &= \sum_{i=1}^{K} \frac{2^{\operatorname{rel}_i} - 1}{\log_2(i + 1)}, \\
\operatorname{IDCG@K} &= \sum_{i=1}^{|\hat{p}^{\text{gt}}|} \frac{2^{\operatorname{rel}_i^{\text{(ideal)}}} - 1}{\log_2(i + 1)}, \\
\operatorname{NDCG@K} &= \frac{\operatorname{DCG@K}}{\operatorname{IDCG@K}},
\end{align}
where $\operatorname{rel}_i$ is the relevance score (e.g., LI score) of the $i$-th retrieved page, and $\operatorname{rel}_i^{\text{(ideal)}}$ is the relevance score of the $i$-th page in the ideal ranking (sorted by relevance in descending order).

\noindent \textbf{Mean Reciprocal Rank (MRR)}: Computes the average of the reciprocal rank of the first relevant page. The calculation formula is as follows:
\begin{align}
\operatorname{MRR} = \frac{1}{Q_D} \sum_{j=1}^{Q_D} \frac{1}{\operatorname{rank}_j},
\end{align}
where $Q_D$ is the number of queries, and $\operatorname{rank}_j$ is the rank of the first relevant page for the $j$-th query.

These metrics collectively provide a comprehensive view of the retrieval process: Recall emphasizes coverage, Precision emphasizes accuracy, and NDCG and MRR emphasize ranking quality.

\begin{figure*}[!t]
\centering
\includegraphics[width=1\linewidth]{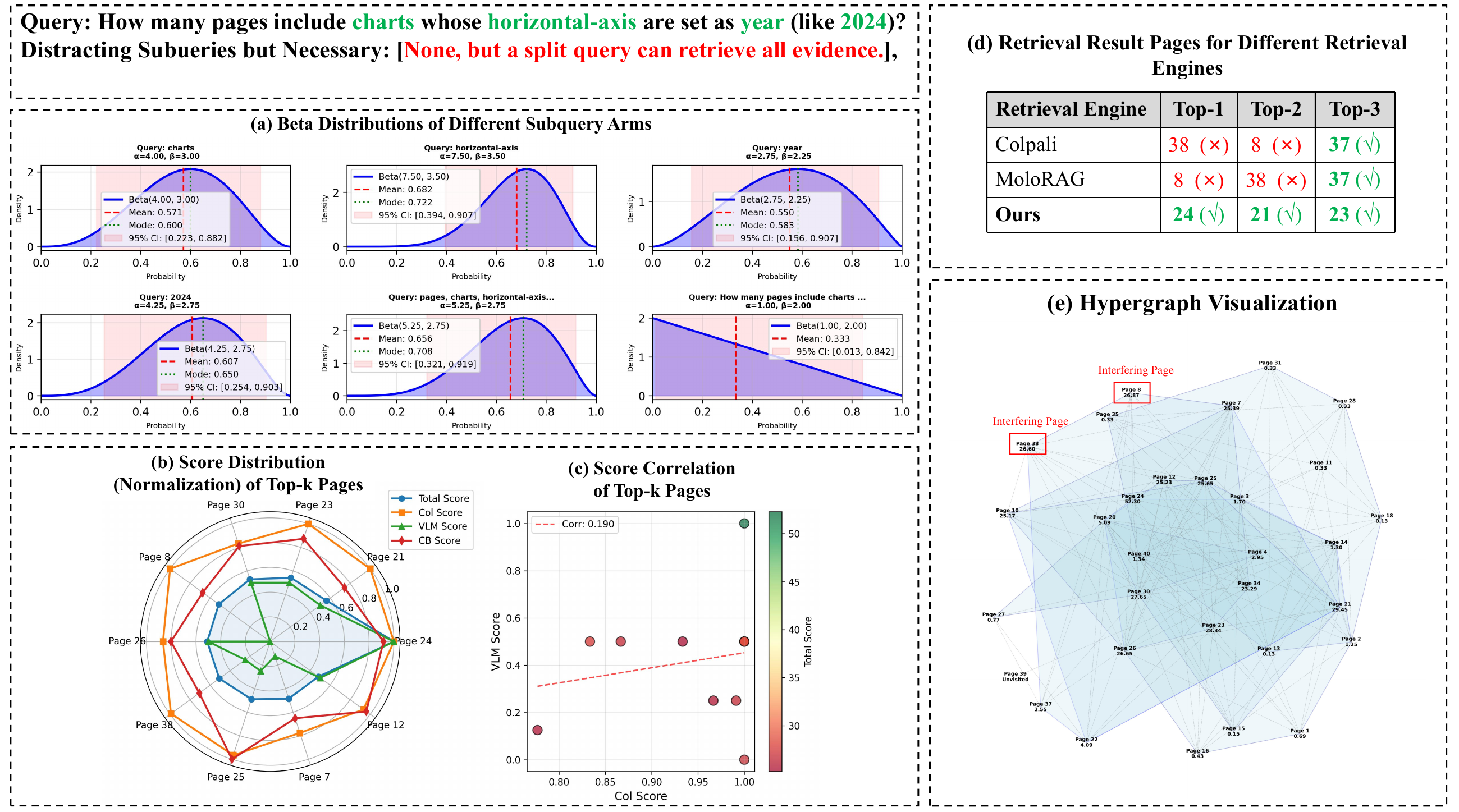}
\caption{Qualitative Analysis of the MAB-DQA Framework.
We selected a representative query that requires extensive page regression.
The evidence for the problem spans 13 pages (specifically required: 10, 12, 14, 15, 20, 21, 22, 23, 24, 25, 26, 30, 37). It is evident that this query contains a large number of correct pages.
This example demonstrates that our method, in the presence of positive samples, recalls a broader range of correct pages more comprehensively.
As shown in Fig.~(a), MAB-DQA, by providing query decomposition, recalls more correct evidence pages.
As shown in Fig.~(e), when extensive recall is required, distractor pages are usually contained within a small number of hyperedges.}
\label{fig:case6}
\end{figure*}

\begin{figure*}[!t]
\centering
\includegraphics[width=1\linewidth]{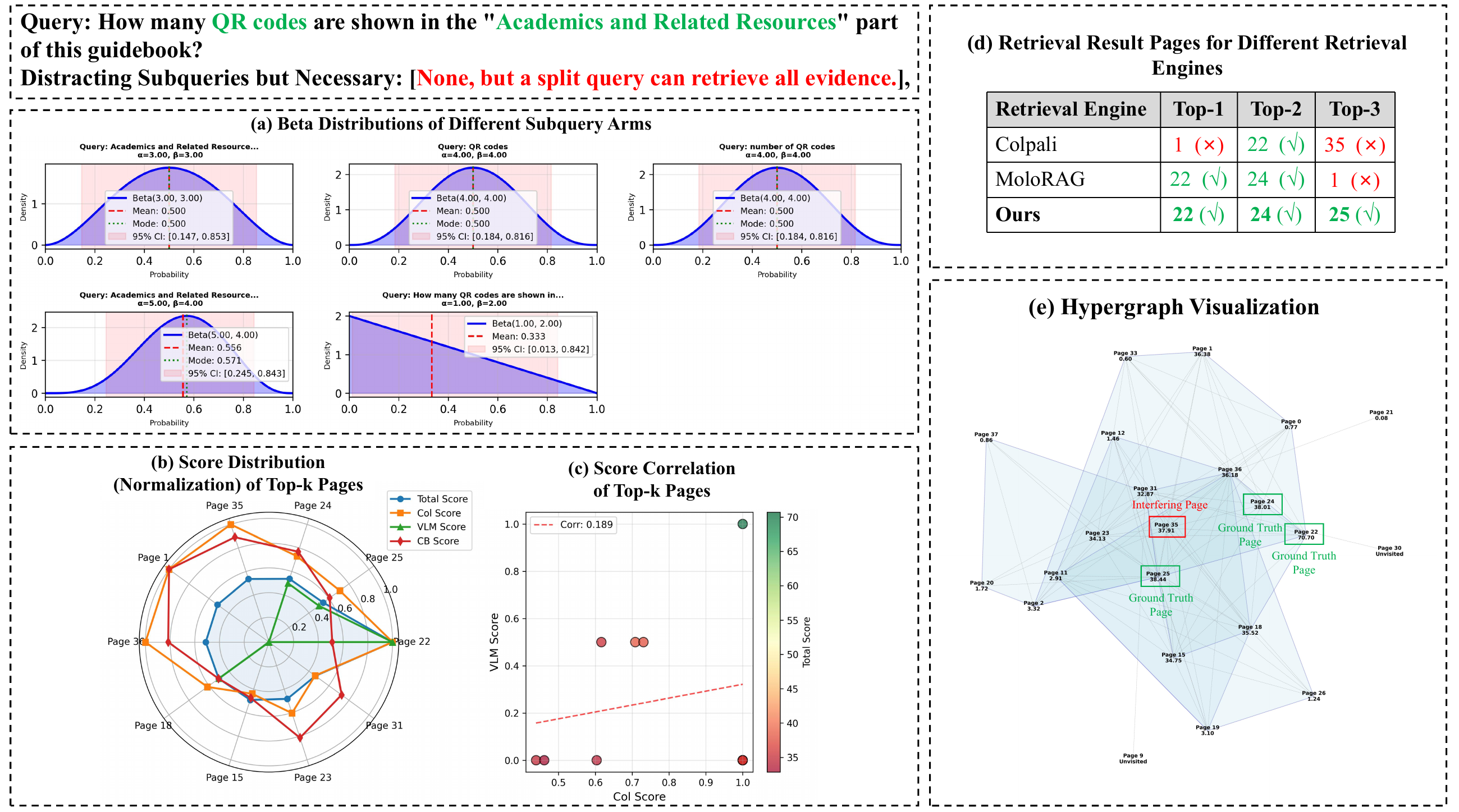}
\caption{Qualitative Analysis of the MAB-DQA Framework.
We selected a representative query that requires a multi-hop answer.
The evidence for the problem spans 3 pages (specifically required: 22, 24, 25).
This query requires the QA framework to both accurately identify evidence and completely recall it.
As shown in Fig.~(b), the initial scores provided by Colpali include distractor pages (35, 1).
As shown in Fig.~(d), through multiple rounds of online evaluation, MAB-DQA correctly excludes the distractor pages.
MAB-DQA provides an online reranking algorithm that correctly associates all evidence.}
\label{fig:case7}
\end{figure*}

\begin{figure*}[!t]
\centering
\includegraphics[width=1\linewidth]{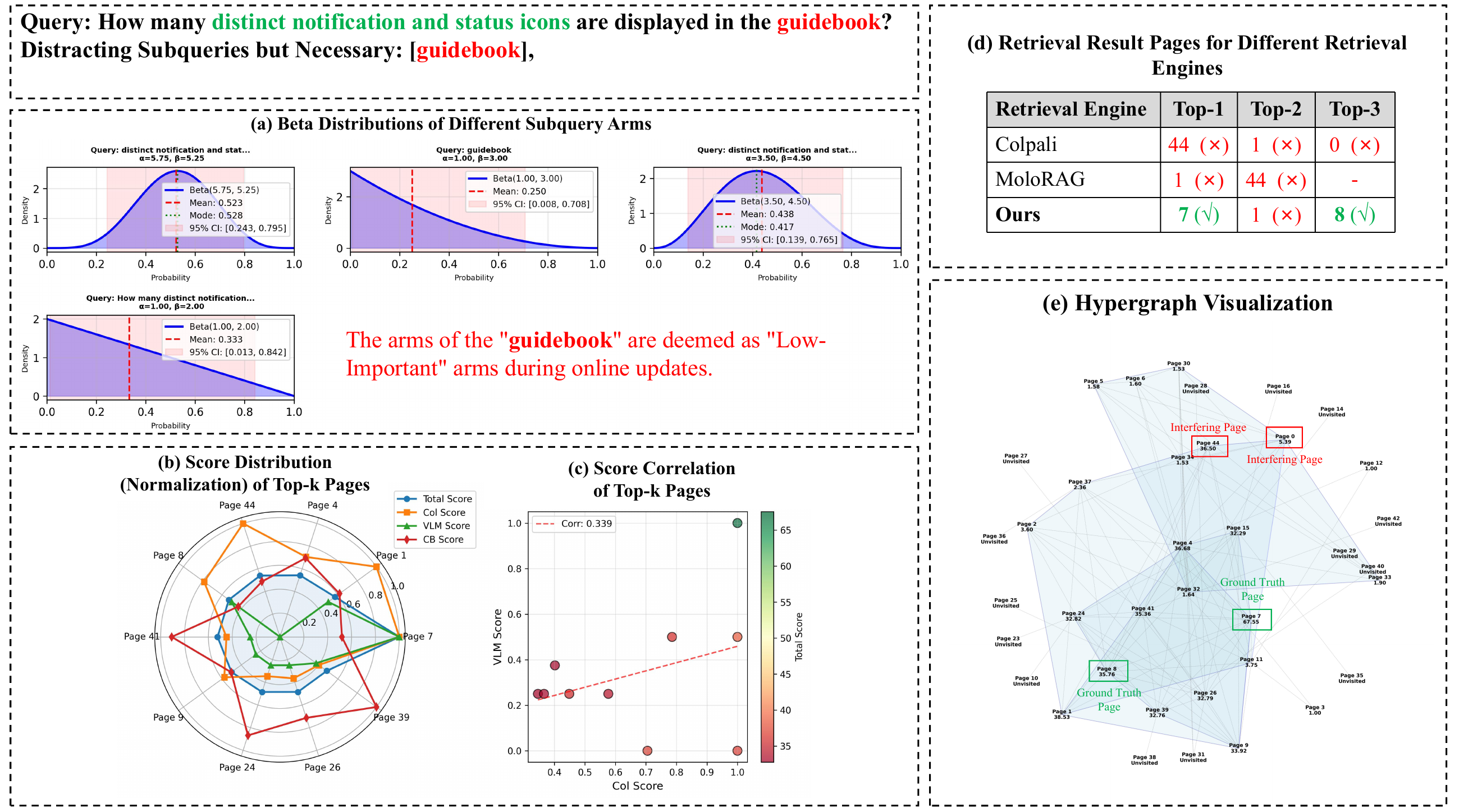}
\caption{Qualitative Analysis of the MAB-DQA Framework.
We selected a representative query that requires avoiding distracting conditions.
The evidence for the problem spans 2 pages (specifically required: 7, 8).
"Guide book" is likely a high-frequency term that appears frequently throughout the entire document.
This causes the retrieval engine to consider every page as directly relevant to the evidence for this problem.
As shown in Fig.~(d), both Colpali and MoloRAG retrieve incorrectly.
MAB-DQA avoids the unimportant condition "guide book," enabling the recall of the correct evidence pages.}
\label{fig:case8}
\end{figure*}

\section{Additional Case Studies and Qualitative Analysis}
\label{sec:appendix_case}

The Appendix~\ref{sec:appendix_case} provides detailed qualitative analyses to further illustrate the effectiveness of the proposed MAB-DQA framework in handling diverse multi-aspect multi-modality DQA scenarios. Through three representative case studies, we demonstrate how MAB-DQA addresses key challenges such as extensive page regression, multi-hop reasoning, and distraction from high-frequency terms. Each case includes a visual breakdown of the retrieval process, highlighting the framework's ability to dynamically weigh query conditions via hypergraph-based retrieval and reflective reasoning. The figures below present real examples from the MMLongBench benchmark, with annotations to clarify the retrieval mechanisms.

In Fig.~\ref{fig:case6}, the query involves evidence spread across 13 pages, testing the framework's ability to handle large-scale regression. MAB-DQA's query decomposition and hypergraph structure allow it to prioritize relevant pages while containing distractors within sparse hyperedges. Fig.~\ref{fig:case7} focuses on a multi-hop query, where the initial retrieval by baselines includes irrelevant pages (e.g., page 35). Through iterative VLM feedback, MAB-DQA refines the path to associate all correct evidence. Fig.~\ref{fig:case8} addresses a common pitfall where high-frequency terms like "guide book" mislead retrieval. By evaluating condition importance online, MAB-DQA suppresses such distractions and accurately locates the two evidence pages. Collectively, these cases validate the framework's robustness in real-world multi-modality DQA settings.

\section{Prompt Settings} 
\label{sec:appendix_prompt}

The Appendix~\ref{sec:appendix_prompt} section provides a comprehensive collection of the prompt engineering templates utilized throughout the MAB-DQA framework. These carefully crafted prompts play a crucial role in enabling the framework's multi-aspect multi-modality importance-aware retrieval and reflective reasoning capabilities. Each prompt serves a specific function in the pipeline.

\subsection{Prompt for Decompose Query}
\label{sec:appendix_decompose}

The following prompt template is designed to extract meaningful subqueries from complex user questions, which form the foundation for the hypergraph-based retrieval approach:

\begin{tcolorbox}
[colback=gray!20, colframe=gray!100, sharp corners, leftrule={3pt}, rightrule={0pt}, toprule={0pt}, bottomrule={0pt}, left={2pt}, right={2pt}, top={3pt}, bottom={3pt}]

As an AI agent specialized in document retrieval query processing, your primary task is to handle each query by first ignoring any irrelevant information (such as output format requests or non-retrieval instructions).

Then, extract meaningful entities and key phrases that capture the core intent of the query.

Finally, output the result as a comma-separated list of key phrases, for example: "key\_phrase1, key\_phrase2, ...". Ensure clarity and conciseness throughout.

\end{tcolorbox}

\subsection{Prompt for Evaluate the Retrieval Evidence}
\label{sec:appendix_retrieval_evidence}

\begin{tcolorbox}
[colback=gray!20, colframe=gray!100, sharp corners, leftrule={3pt}, rightrule={0pt}, toprule={0pt}, bottomrule={0pt}, left={2pt}, right={2pt}, top={3pt}, bottom={3pt}, breakable,]

\# GOAL \# You are a Retrieval Expert, and your task is to evaluate how relevant the input document page is to the given query. 

Rate the relevance on a scale of 1 to 5, where: 

- 5: Highly relevant - contains COMPLETE information to fully answer the query (be cautious with this rating) 

- 4: Very relevant - contains most information needed but may lack some details (be cautious with this rating) 

- 3: Moderately relevant - contains some useful information but significant gaps remain 

- 2: Slightly relevant - has minor connection to the query 

- 1: Irrelevant - contains no information related to the query 

\# INSTRUCTION \# Based on previous retrieval system judgment, we believe that this document snapshot is at least ' + priori + ' relevant. Please first read the given query, think about what specific information is required to answer that query comprehensively, and then carefully examine the document snapshot. 

\# IMPORTANT \# Before giving a score of 4 or 5, verify that the page actually contains the specific facts needed to answer the query, not just related information. 

\# QUERY\# + query + Think step by step about the relevance, then provide just a single number (1-5) representing your judgment.

\end{tcolorbox}

\subsection{Prompt For Evaluation:}

\begin{tcolorbox}
[colback=gray!20, colframe=gray!100, sharp corners, leftrule={3pt}, rightrule={0pt}, toprule={0pt}, bottomrule={0pt}, left={2pt}, right={2pt}, top={3pt}, bottom={3pt}, breakable,]

Question: \{question\}

Predicted Answer: \{answer\}

Ground Truth Answer: \{gt\}

Please evaluate if the predicted answer is correct compared to the ground truth, considering the following criteria:

- If the Ground Truth Answer is "Not answerable":

- And the Predicted Answer indicates that the model cannot answer, then it is considered CORRECT (score 1).
  
- Otherwise:

- Score based on whether the Predicted Answer is factually and logically consistent with the Ground Truth Answer.
  
Score the answer on Binary correctness (0-1): 1 if the answer is correct, 0 if it is incorrect
Return only a JSON-parsable string in the format: 
\{\{"binary\_correctness": <score>\}\}

Output:

\end{tcolorbox}

\subsection{Prompt For Question Answering}

The evaluation prompt ensures consistent assessment of answer quality across different benchmarks, maintaining standardization in the baseline performance measurement:

\begin{tcolorbox}
[colback=gray!20, colframe=gray!100, sharp corners, leftrule={3pt}, rightrule={0pt}, toprule={0pt}, bottomrule={0pt}, left={2pt}, right={2pt}, top={3pt}, bottom={3pt}, breakable,]

Using the provided \{ num\_images \} document screenshots, answer this question: "\{question\}"

Requirements:

- Reply must be extremely concise (as short as possible)

- Use only information visible in the screenshots

- If the answer cannot be clearly found, respond exactly: "Not answerable"

Answer:

\end{tcolorbox}

\subsection{Prompt For Question Reflection}

The reflection prompt enables the HRRA component to refine ambiguous queries, improving retrieval precision through iterative clarification:

\begin{tcolorbox}
[colback=gray!20, colframe=gray!100, sharp corners, leftrule={3pt}, rightrule={0pt}, toprule={0pt}, bottomrule={0pt}, left={2pt}, right={2pt}, top={3pt}, bottom={3pt}, breakable,]

Based on the provided \{ num\_images \} document screenshots, rephrase the following question to make it clearer and more specific.

Original question: "\{ question \}"

Requirements for rewriting:

1. If the question is clear and can be answered using ONLY information in the screenshots, keep it essentially the same

2. If the question is ambiguous or vague, clarify it based on what information appears to be available in the screenshots

3. If the question cannot be answered with the screenshots, note this, but still try to rephrase for clarity

4. The rewritten question should be specific, direct, and answerable using visible document content

5. Keep the core intent of the original question

6. If screenshots show specific entities (names, dates, numbers, terms), use them in the rewritten question

7. Output only the rewritten question, nothing else

Rewritten question:

\end{tcolorbox}

\subsection{Prompt For Answer Reflection}

This prompt template facilitates the multi-stage verification process by assessing whether answers adequately address the original query requirements:

\begin{tcolorbox}
[colback=gray!20, colframe=gray!100, sharp corners, leftrule={3pt}, rightrule={0pt}, toprule={0pt}, bottomrule={0pt}, left={2pt}, right={2pt}, top={3pt}, bottom={3pt}, breakable,]

You will be given a question and a corresponding answer. Your task is to determine whether the answer addresses the question, regardless of whether the answer is correct or not.

Focus only on whether the answer responds to the question and covers the necessary points (i.e., no essential content is missing).

If no answer is provided, consider it as not answering.

Question: \{ question \}

Answer: \{ answer \}

Did the answer address the question? (yes/no)

\end{tcolorbox}

\subsection{Prompt For Hypergraph Summary}

The hypergraph summary prompt enables structural analysis of queries, 
summarizing their associations from hypergraph, and supporting the framework's ability to handle multi-hop reasoning tasks:

\begin{tcolorbox}
[colback=gray!20, colframe=gray!100, sharp corners, leftrule={3pt}, rightrule={0pt}, toprule={0pt}, bottomrule={0pt}, left={2pt}, right={2pt}, top={3pt}, bottom={3pt}, breakable,]

Analyze the following question and identify the core concepts and relationships that need to be understood to answer it properly.

Question: "\{ question \}"
\{ "Key aspects to focus on: " + hypergraph if hypergraph else "Identify the key concepts and relationships in this question." \}

Requirements:

- Break down the question into fundamental components

- Identify what specific information is needed to answer each component

- Note any implicit relationships or assumptions in the question

- Be concise but thorough in your analysis

Analysis:

\end{tcolorbox}

\subsection{Prompt For Refined Question Answering}

The final refinement prompt implements the critical thinking component of the HRRA, enabling iterative improvement of initial answers through evidence synthesis and reflection:

\begin{tcolorbox}
[colback=gray!20, colframe=gray!100, sharp corners, leftrule={3pt}, rightrule={0pt}, toprule={0pt}, bottomrule={0pt}, left={2pt}, right={2pt}, top={3pt}, bottom={3pt}, breakable,]

Based on the following context, provide a better answer to the question through careful reasoning.

Question: \{question\}

Initial incomplete answer: \{initial\_answer\}

Relevant information summary: \{summary\}

CRITICAL THINKING REQUIREMENTS:

1. First, analyze what the question is REALLY asking for

2. Compare the initial answer with the available information

3. Identify gaps or inaccuracies in the initial answer

4. Synthesize information from the summary to fill these gaps

5. Formulate a coherent response that directly addresses the question

DO NOT simply copy phrases from the summary. Instead, use the information to construct a thoughtful answer.

If the summary indicates no relevant information, respond: "Not answerable"

Reasoning process:

- [Analyze the question requirements]

- [Compare initial answer with evidence]

- [Identify what needs to be improved]

- [Synthesize the improved answer]

Improved answer:

\end{tcolorbox}

\section{Algorithm Design} 
\label{sec:appendix_algorithm}

\begin{algorithm*}[!t]
\caption{Query-Aware Page Hypergraph Construction}
\label{alg:hypergraph-construction}
\normalsize
\begin{algorithmic}[1]
\Require Query $q$, document pages $\{p_1, p_2, \dots, p_N\}$, 
        VLM $\mathcal{V}$, graph threshold $\theta_{\mathbf{G}}$, hyperedge capacity $\theta_{\mathbf{H}}$
\Ensure Hypergraph $\mathbf{H} = (\mathbf{V}_{\mathbf{G}}, \mathbf{E}_{\mathbf{H}} \cup \mathbf{E}_{\mathbf{G}})$
\Statex
\Procedure{HypergraphConstruction}{$q, \{p_i\}_{i=1}^N, \theta_{\mathbf{G}}, \theta_{\mathbf{H}}$}
    \Comment{Construct hypergraph structure from documents and query}
    \State $\mathbf{E}_{\mathbf{q}}, \mathbf{E}_{\mathbf{p}_i} \gets$ VLM embeddings of $q$ and $p_i$
    \State $\mathbf{G}(\mathbf{V}_{\mathbf{G}}, \mathbf{E}_{\mathbf{G}}) \gets$ empty graph
    \For{$i = 1$ to $N$}
        \For{$j = i+1$ to $N$}
            \If{$\simop(\mathbf{E}_{p_i}, \mathbf{E}_{p_j}) \geq \theta_{\mathbf{G}}$}
                \State $\mathbf{E}_{\mathbf{G}} \gets \mathbf{E}_{\mathbf{G}} \cup \{ \{p_i, p_j\} \}$
            \EndIf
        \EndFor
    \EndFor
    \State $\mathcal{E}_q \gets \{q_1, q_2, \dots, q_M\} \gets \mathcal{V}.\text{decompose}(q)$ \Comment{VLM decomposes query}
    \State $\mathbf{Q} \gets \{ \{\hat{q}\} \mid \hat{q} \in \mathcal{E}_q \} \cup \{ \mathcal{E}_q \}$ \Comment{Atom‑Integral Subqueries Set}
    \State Let $\mathbf{Q}_b = \mathcal{E}_q$ (global reference, $b = M+1$)
    \State $\mathbf{H} \gets (\mathbf{V}_{\mathbf{G}}, \mathbf{E}_{\mathbf{H}} \cup \mathbf{E}_{\mathbf{G}})$ \Comment{Composite hypergraph}
    \For{$j = 1$ to $M+1$}
        \State $\mathbf{C}_j \gets \text{Top-$\theta_{\mathbf{H}}$ pages by } \LI(\mathbf{Q}_j, p)$
        \Comment{Select top pages for each query subset}
        \State $\hat{\mathbf{E}}_j \gets \{ p \in \mathbf{C}_j \mid p \notin \mathbf{C}_b \vee \rank(\LI(\mathbf{Q}_j, p)) \leq \rank(\LI(\mathbf{Q}_b, p)) \}$
        \Comment{Filter pages based on global reference}
        \State $\mathbf{E}_{\mathbf{H}} \gets \mathbf{E}_{\mathbf{H}} \cup \{ \hat{\mathbf{E}}_j \}$
    \EndFor
    \State \Return $\mathbf{H}$
\EndProcedure
\end{algorithmic}
\end{algorithm*}

\begin{algorithm*}[!t]
\caption{Multi-Armed Bandit–based Retrieval}
\label{alg:mabr}
\normalsize
\begin{algorithmic}[1]
\Require Hypergraph $\mathbf{H} = (\mathbf{V}_{\mathbf{G}}, \mathbf{E}_{\mathbf{H}} \cup \mathbf{E}_{\mathbf{G}})$, 
        retrieval iteration $m$, hyperparameters $\alpha, \beta, \lambda$, 
        VLM $\mathcal{V}$, query $q$, original pages $\{p_i\}_{i=1}^N$
\Ensure Top-10 retrieved page indices $\mathbf{R}_{\text{top10}}$
\Statex
\Procedure{MAB-Retrieval}{$\mathbf{H}, m, \alpha, \beta, \lambda$}
    \Comment{Hypergraph Bandit Thompson Sampling Search}
    \State Initialize bandit arms: $\forall j, \alpha_j = 1, \beta_j = 1$
    \State $\text{visited} \gets \emptyset$, $\text{scores} \gets \emptyset$, $\text{current\_page} \gets \emptyset$
    \State Initialize $\text{score}(p_i) = \mathbf{LI}(q,p_i)$ for all $p_i \in \mathbf{V}_{\mathbf{H}}$
    \For{iteration $t = 1$ to $m$}
        \Comment{Main search loop}
        \State $\text{to\_evaluate} \gets \emptyset$  \Comment{Set of pages to evaluate in this iteration}
        
        \If{$\text{current\_page} = \emptyset$}
            \Comment{No current page, evaluate all pages}
            \State $\text{to\_evaluate} \gets \mathbf{V}_{\mathbf{H}}$
        \Else
            \Comment{Start from current page}
            \State $\text{to\_evaluate} \gets \{\text{current\_page}\}$
        \EndIf
        
        \For{each page $p_i \in \text{to\_evaluate}$}
            \State $\hat{\mathbf{Q}}_i \gets$ subquery linked to $p_i$
            \State $s_i^{\text{cb}} \gets \frac{1}{|\hat{\mathbf{Q}}_i|} \sum_{\mathbf{Q}_j \in \hat{\mathbf{Q}}_i} \mathbb{E}[\Beta(\alpha_j, \beta_j)]$
            \Comment{Bandit score from Thompson sampling}
            \State $h_i \gets \deg(p_i) \text{ in } \mathbf{H}$
            
            \If{$p_i \in \text{visited}$}
                \State $\text{Adj}_{\mathbf{H}}(p_i) \gets \{p_j \in \mathbf{V}_{\mathbf{H}} \mid \exists e \in \mathbf{E}_{\mathbf{H}}, \{p_i, p_j\} \subseteq e\}$
                \State $\text{Unvisited} \gets \text{Adj}_{\mathbf{H}}(p_i) \setminus \text{visited}$
                \If{$\text{Unvisited} \neq \emptyset$}
                    \Comment{Found unvisited neighbors, jump to the best one}
                    \State $p^* \gets \arg\max_{p_j \in \text{Unvisited}} \text{score}(p_j)$
                    \State $\text{current\_page} \gets p^*$  \Comment{Update current page for next iteration}
                \Else
                    \State Continue  \Comment{No unvisited neighbors, reset}
                \EndIf
            \Else
                \State $s_i^{\text{vlm}} \gets$ VLM evaluates relevance of $p_i$ to $q$
                \Comment{VLM evaluation for unvisited nodes}
                \State $\text{visited} \gets \text{visited} \cup \{p_i\}$
                \State $\text{current\_page} \gets p_i$  \Comment{Stay on this page for potential jump next iteration}
            \EndIf
            
            \State $\text{score}(p_i) \gets (1-\alpha)\max_j \LI(\mathbf{Q}_j, p_i) + \alpha s_i^{\text{vlm}} + \beta[(1-\lambda)h_i + \lambda s_i^{\text{cb}}]$
            \Comment{Composite scoring function}
            \State $\text{scores}[p_i] \gets \text{score}(p_i)$
        \EndFor
        
        \Comment{Update bandit parameters for all visited pages}
        \For{each $p_i \in \text{visited}$}
            \State $\hat{\mathbf{Q}}_i \gets$ linked subqueries
            \If{$p_i$ was evaluated by VLM in this iteration}
                \State $\forall \mathbf{Q}_j \in \hat{\mathbf{Q}}_i: (\alpha_j, \beta_j) \gets (\alpha_j + s_i^{\text{vlm}}, \beta_j + 1 - s_i^{\text{vlm}})$
            \EndIf
        \EndFor
    \EndFor
    \State $\mathbf{R} \gets \text{top } 10 \text{ pages by final scores}$
    \State \Return $\mathbf{R}$
\EndProcedure
\end{algorithmic}
\end{algorithm*}

This section provides detailed algorithmic descriptions of the two core components in our MAB-DQA framework: the hypergraph construction process and the MABR algorithm.

\subsection{Hypergraph Construction Algorithm}

As shown in Algorithm~1, the hypergraph construction algorithm serves as the foundation for our multi-aspect multi-modality retrieval framework. It transforms the original document and query into a structured hypergraph representation that captures complex conditional associations.

The algorithm begins by computing visual-language embeddings for both the query and all document pages. It then constructs a query-agnostic page graph $\mathbf{G}$ where edges connect pages with similarity scores exceeding the threshold $\theta_\mathbf{G}$. This graph captures the intrinsic relationships between pages based on their semantic content.

A key innovation is the decomposition of the original query $q$ into subqueries ${q_1,q_2,\dots,q_M}$ using the VLM, forming an Atomic and Global Subqueries Set that includes both individual subqueries and their complete combination. For each query subset $\mathbf{Q}_j$, the algorithm selects the top-$\theta_\mathbf{H}$ pages based on late interaction scores, then filters them against a global reference set $\mathbf{Q}_b$ to ensure that only pages with improved ranking under the specific subquery are included in the hyperedge.

The time complexity of Algorithm~\ref{alg:hypergraph-construction} is $O(N^2\cdot D+M\cdot N\cdot D)$, where $N$ is number of pages, $M$ is number of subqueries, and $D$ is embedding dimension. The quadratic term arises from the query-agnostic page graph construction, while the linear term accounts for the hyperedge generation.

\subsection{Multi-Armed Bandit-based Retrieval}

Algorithm~\ref{alg:mabr} implements our novel retrieval strategy that formulates the retrieval process as a combinatorial multi-armed bandit problem. Each subquery $\mathbf{Q}_j$ is treated as an "Arm" with a Beta distribution $\text{Beta}(\alpha_j,\beta_j)$ modeling its reward distribution.

The algorithm maintains a dynamic scoring function that combines four components: (1) a late interaction score between the page and any subquery; (2) a direct VLM relevance assessment $s_i^{vlm}$; (3) a hypergraph-based term balancing page connectivity $h_i$; (4) a bandit confidence scores $s_i^{cb}$.

The retrieval proceeds iteratively, with the beam focusing on the most promising pages based on the composite score. Thompson sampling ensures a balance between exploration (trying less certain subqueries) and exploitation (focusing on combinations that have yielded high rewards). After each VLM evaluation, the algorithm updates the Beta parameters for all subqueries associated with the evaluated page, enabling online learning of the query's condition importance.

Ultimately, the time complexity of Algorithm~\ref{alg:mabr} is $O(m\cdot T_{\text{VLM}} +m\cdot |E_\mathbf{H}|)$, where $m$ is the number of iterations and $T_{\text{VLM}}$ is the VLM evaluation time. The MABR algorithm's efficiency stems from its focused preliminary evaluation strategy that prioritizes pages with high potential relevance while maintaining theoretical guarantees through the multi-armed bandit formulation.

\section{Randomness Statement and Control Measures}
\label{sec:appendix_random}

In the experiments presented in this paper, randomness primarily stems from the following sources:
(1) The Visual Language Model (VLM) for query decomposition may generate different subqueries; the VLM temperature is fixed at 0, but there may still be inherent stochastic risks.
(2) The random initialization in page similarity computations (e.g., embedding generation) during the hypergraph construction process.
(3) Uncertainties arising during hyperedge construction.
(4) The random exploration strategy of Thompson Sampling in the multi-armed bandit problem.
These stochastic factors may cause slight fluctuations in the results of a single experimental run. To mitigate their impact, we ensure that all key experiments are conducted with multiple independent runs (the specific number is detailed in the experimental setup) and employ a fixed random seed (set to 42) to guarantee reproducibility. For instance, in the hyperparameter sensitivity analysis (Sec.~\ref{sec:sensitivity}), we repeat experiments using a controlled variable method to isolate random noise. All experimental results reported in this paper are based on multiple runs (typically 3), and the best performance values are reported to demonstrate the potential of the method.
In our experiments, we observe that performance fluctuations due to randomness are approximately $\pm 0.5\%$, which is considered acceptable. This assessment is primarily based on the following reasons: (1) Sensitivity analysis (Sec.~4.4) shows that variations in hyperparameters (such as $\alpha, \beta, \lambda$) exhibit distinct peaks in their impact on the metrics; (2) Results across multiple datasets (e.g., MMLongBench, LongDocURL with manually labeled retrieval tags) are highly consistent, and the observed fluctuations do not significantly alter the conclusions.

\begin{table*}[ht]
\caption{Individual Component Ablation Study on MAB-DQA Framework}
\centering
\small
\begin{tabular}{lcccc}
\toprule
\textbf{Model Variant} &
\textbf{MMLongBench} & \textbf{LongDocURL} & \textbf{FetaTab} & \textbf{PaperTab}\\
\midrule

w/o Query-Agnostic Page Graph & 
0.394 & 0.560 & 0.636 & 0.247 \\

w/o Atomic and Global Subqueries Set &
0.317 & 0.482 & 0.543 & 0.198 \\

w/o Question Reflection &
\textbf{0.401} & 0.556 & 0.631 & 0.219 \\

w/o Answer Reflection &
0.397 & 0.561 & 0.625 & 0.247 \\

\midrule

Ours (Full Model) &
0.399 & \textbf{0.564} & \textbf{0.638} & \textbf{0.269}\\

\bottomrule
\end{tabular}
\label{tab:appendix_ablation}
\end{table*}

\section{Supplementary Ablation and Sensitivity Experiments}
\label{sec:appendix_experiment}

\begin{figure}[!t]
\centering
\includegraphics[width=1.\linewidth]{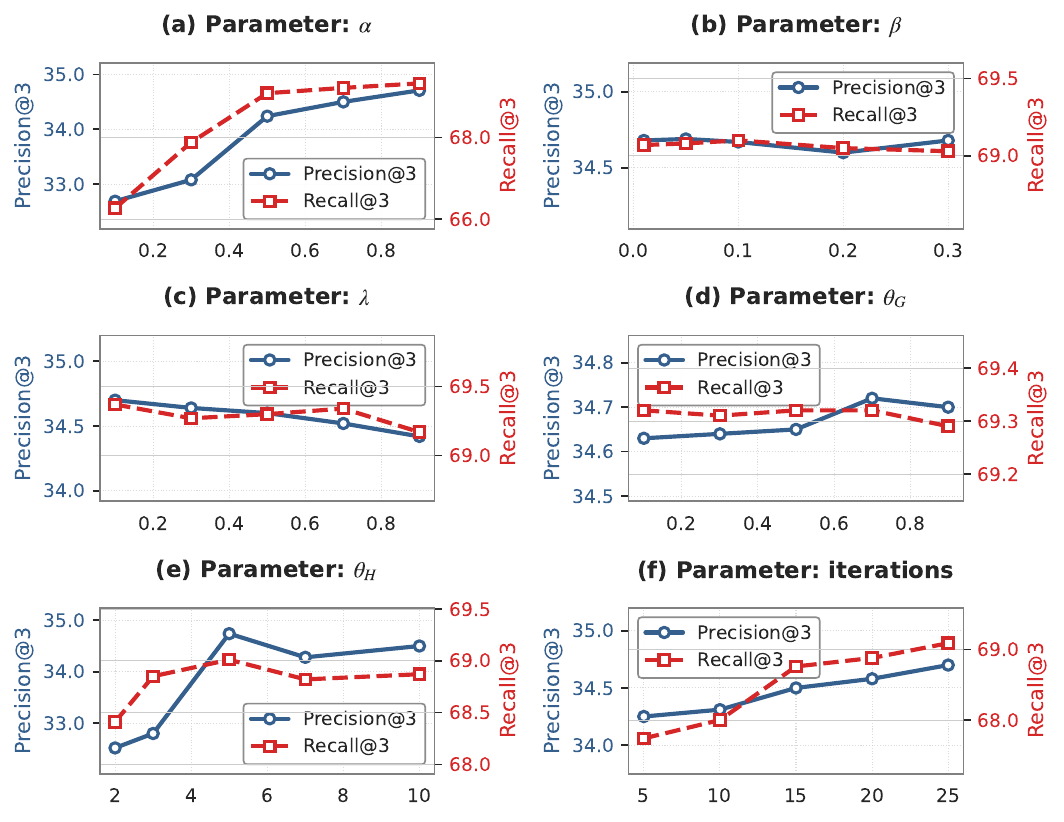}
\caption{Sensitivity Analysis of Key Hyperparameters of the MAB-DQA Framework under Top-3 Retrieval on MMLongBench. The blue dashed line represents Precision. The red line represents Recall.}
\label{fig:appendix_sensitivity}
\end{figure}

The Appendix~\ref{sec:appendix_experiment} presents additional ablation studies that investigate the individual contributions of key components in the MAB-DQA framework. Unlike the progressive ablation approach in the main text (Sec.~\ref{sec:ablation}), which sequentially added modules, here we examine the impact of removing single components while keeping others intact. This provides a more granular understanding of each module's role in the overall system performance.
All experiments maintained the same evaluation metrics (accuracy) and experimental conditions as described in Sec.~4.1. In Fig.~\ref{fig:appendix_sensitivity}, we have supplemented the sensitivity analysis of the MAB-DQA framework on the MMLongBench dataset. Table~\ref{tab:appendix_ablation} presents the quantitative results of the individual component ablation study.

The ablation experiments were conducted on the same four open-source benchmark datasets as in the main experiments: MMLongBench, LongDocURL, FetaTab, and PaperTab. We evaluated four modified versions of our framework by individually removing specific components:

\begin{itemize}
\item \textbf{w/o Query-Agnostic Page Graph}: Removes the Query-Agnostic Page Graph construction (as shown in Sec.~3.1), disabling the modeling of inter-page relationships.

\item \textbf{w/o Atomic and Global Subqueries Set
}: Uses only the Atomic Set (as shown in Eq.~3), limiting the framework's ability to capture global information.

\item \textbf{w/o Question Reflection}: Disable the question reflection component in HRRA, thus eliminating the entire query clarification and refinement process.

\item \textbf{w/o Answer Reflection}: Removes the answer reflection mechanism in the HRRA, disabling the multi-stage verification and refinement of generated answers.
\end{itemize}

From the above analysis, several key observations emerge:

\textbf{Query-Agnostic Page Graph Contribution}: The removal of the Query-Agnostic Page Graph Contribution causes performance degradation across all datasets, with the most significant impact on PaperTab (-8.2\%), which contains complex tabular structures. This demonstrates that modeling inter-page relationships is particularly crucial for documents with strong structural dependencies.

\textbf{Atomic and Global Subqueries Set}: Removing query decomposition results in the most substantial performance drop overall, particularly on LongDocURL (-14.5\%), which contains complex multi-aspect questions. This highlights that capturing both fine-grained and holistic query aspects is essential for handling diverse question types.

\textbf{Reflection Components}: The question and answer reflection mechanisms exhibit clear complementary effects with each other.

The impact of individual components varies across datasets. PaperTab, focusing on scientific document understanding, benefits most from structural components (similarity graph), while LongDocURL, emphasizing long-document reasoning, relies heavily on query decomposition strategies. These findings confirm that each component in MAB-DQA contributes uniquely to the framework's overall effectiveness, with different modules playing dominant roles depending on the specific document characteristics and question types encountered in the DQA task.

\end{document}